\documentclass[10pt,conference]{IEEEtran}
\IEEEoverridecommandlockouts

\usepackage{cite}
\usepackage{amsmath,amssymb,amsfonts}
\usepackage{algorithmic}
\usepackage{graphicx}
\usepackage{textcomp}
\usepackage{xcolor}
\def\BibTeX{{\rm B\kern-.05em{\sc i\kern-.025em b}\kern-.08em
    T\kern-.1667em\lower.7ex\hbox{E}\kern-.125emX}}

\usepackage[hidelinks]{hyperref}
\usepackage{bm}
\usepackage{comment}
\usepackage{algorithm}
\usepackage{url}
\usepackage{mathtools}

\newcommand{\argmin}{\mathop{\rm arg~min}\limits}

\begin{document}

\title{Self-Organizing Maps with\\Optimized Latent Positions\\
\thanks{This work was supported by JSPS KAKENHI Grant Number JP24K15110.}
\thanks{\copyright\ 2026 IEEE. Personal use of this material is permitted. Permission from IEEE must be obtained for all other uses, in any current or future media, including reprinting/republishing this material for advertising or promotional purposes, creating new collective works, for resale or redistribution to servers or lists, or reuse of any copyrighted component of this work in other works.}
\thanks{This is the accepted version of the article: S. Ubukata, A. Notsu, and K. Honda, ``Self-Organizing Maps with Optimized Latent Positions,'' in 2026 International Joint Conference on Neural Networks (IJCNN), 2026, pp.~1--8, doi: \href{https://doi.org/10.1109/ijcnn69854.2026.11729950}{10.1109/ijcnn69854.2026.11729950}.}}

\author{\IEEEauthorblockN{1\textsuperscript{st} Seiki Ubukata}
    \IEEEauthorblockA{\textit{Graduate School of Informatics} \\
        \textit{Osaka Metropolitan University}\\
        Osaka 599-8531, Japan \\
        ubukata@omu.ac.jp}
    \and
    \IEEEauthorblockN{2\textsuperscript{nd} Akira Notsu}
    \IEEEauthorblockA{\textit{Graduate School of Sustainable System Sciences} \\
        \textit{Osaka Metropolitan University}\\
        Osaka 599-8531, Japan \\
        notsu@omu.ac.jp}
    \and
    \IEEEauthorblockN{3\textsuperscript{rd} Katsuhiro Honda}
    \IEEEauthorblockA{\textit{Graduate School of Informatics} \\
        \textit{Osaka Metropolitan University}\\
        Osaka 599-8531, Japan \\
        khonda@omu.ac.jp}
}

\maketitle


\begin{abstract}
Self-Organizing Maps (SOM) are a classical method for unsupervised learning, vector quantization, and topographic mapping of high-dimensional data. However, existing SOM formulations often involve a trade-off between computational efficiency and a clearly defined optimization objective. Objective-based variants such as Soft Topographic Vector Quantization (STVQ) provide a principled formulation, but their neighborhood-coupled computations become expensive as the number of latent nodes increases. In this paper, we propose Self-Organizing Maps with Optimized Latent Positions (SOM-OLP), an objective-based topographic mapping method that introduces a continuous latent position for each data point. Starting from the neighborhood distortion of STVQ, we construct a separable surrogate local cost based on its local quadratic structure and formulate an entropy-regularized objective based on it. This yields a simple block coordinate descent scheme with closed-form updates for assignment probabilities, latent positions, and reference vectors, while guaranteeing monotonic non-increase of the objective and retaining linear per-iteration complexity in the numbers of data points and latent nodes. Experiments on a synthetic saddle manifold, scalability studies on the Digits and MNIST datasets, and 16 benchmark datasets show that SOM-OLP achieves competitive neighborhood preservation and quantization performance, favorable scalability for large numbers of latent nodes and large datasets, and the best average rank among the compared methods on the benchmark datasets.
\end{abstract}


\section{Introduction}
\label{sec:introduction}

Self-Organizing Maps (SOM), introduced by Kohonen \cite{Kohonen1982}, are a classical method for unsupervised learning, vector quantization, and topographic mapping of high-dimensional data. Their main appeal lies in organizing reference vectors on a predefined latent topology while preserving neighborhood structure to a useful extent. Owing to this property, SOM and its variants have been widely used in data analysis and visualization. Despite their usefulness, SOM-based methods involve a trade-off between computational efficiency and an explicit objective-based formulation. The original SOM and Batch SOM (BSOM) \cite{Kohonen1990} are efficient and widely used, but their updates are not generally derived from a single explicit smooth objective \cite{Erwin1992}. In contrast, objective-based methods such as convolved-distortion models \cite{Luttrell1994,Heskes1999} and Soft Topographic Vector Quantization (STVQ) \cite{Graepel1997} offer a clearer optimization perspective, but their explicit neighborhood interactions incur an \(O(NM^2)\) per-iteration cost, where \(N\) and \(M\) are the numbers of data points and map nodes, respectively. Although decomposition techniques can reduce the computational cost~\cite{Heskes2001}, neighborhood coupling among latent nodes remains, and each data point is still represented only through assignment weights over predefined discrete nodes.

In this paper, we propose Self-Organizing Maps with Optimized Latent Positions (SOM-OLP), an objective-based topographic mapping method that introduces a continuous latent position for each data point. Starting from the neighborhood distortion of STVQ, we construct a separable surrogate local cost motivated by its local quadratic structure and formulate an entropy-regularized objective based on it. This yields a simple cyclic block coordinate descent (BCD) scheme with closed-form updates for assignment probabilities, latent positions, and reference vectors. As a result, SOM-OLP retains $O(NM)$ per-iteration complexity while providing a continuous latent representation. 

The proposed formulation has two main advantages. First, it provides an explicit objective with monotonic non-increase under cyclic BCD. Second, it introduces continuous latent positions while retaining closed-form updates and linear per-iteration complexity without explicit node--node coupling. Experiments on a synthetic saddle manifold, scalability studies on the Digits and MNIST datasets, and 16 benchmark datasets show that SOM-OLP achieves strong neighborhood preservation and quantization performance, favorable scalability for large numbers of latent nodes and large datasets, and the best average rank among the compared methods on the benchmark datasets. A minimal implementation of SOM-OLP is available at \url{https://github.com/subukata/som-olp} and on Zenodo (\url{https://doi.org/10.5281/zenodo.19547951}).

\section{Related Work}
\label{sec:related_work}

Research on SOM has evolved from heuristic update rules to optimization-based and entropy-regularized formulations. In this paper, we focus on optimization-based SOM-type methods and review their representative formulations and limitations after introducing the notation.

\subsection{Notation and Problem Setting}

Let $\mathcal{X}=\{\bm{x}_i\}_{i=1}^N \subset \mathbb{R}^D$ be a dataset of $N$ data points in the data space $\mathbb{R}^D$, and let $\mathbf{X}=[\bm{x}_1^\top,\dots,\bm{x}_N^\top]^\top \in \mathbb{R}^{N\times D}$ be the corresponding data matrix. SOM-type methods represent these data points by $M$ reference vectors arranged on a predefined topology in an $L$-dimensional latent space $\mathbb{R}^L$. Each node $j \in \{1,\dots,M\}$ has a fixed node coordinate $\bm{r}_j \in \mathbb{R}^L$ and a reference vector $\bm{w}_j \in \mathbb{R}^D$. We denote the sets of node coordinates and reference vectors by $\mathcal{R}=\{\bm{r}_j\}_{j=1}^M$ and $\mathcal{W}=\{\bm{w}_j\}_{j=1}^M$, and the corresponding matrices by $\mathbf{R}=[\bm{r}_1^\top,\dots,\bm{r}_M^\top]^\top \in \mathbb{R}^{M\times L}$ and $\mathbf{W}=[\bm{w}_1^\top,\dots,\bm{w}_M^\top]^\top \in \mathbb{R}^{M\times D}$. When soft assignments are used, the correspondence between data points and nodes is represented by $\mathbf{P}=[p_{ij}] \in [0,1]^{N\times M}$ with $\sum_{j=1}^M p_{ij}=1$ for each data point $i$, where hard assignments are included as a special case.

\subsection{Classical SOM-Type Methods Without a Single Objective Function}

The standard sequential SOM, introduced by Kohonen \cite{Kohonen1982}, assigns each input $\bm{x}_i$ to its best matching unit (BMU),
\begin{equation}
    k_i^* = \argmin_{k \in \{1,\dots,M\}} \|\bm{x}_i - \bm{w}_k\|^2.
\end{equation}
With the neighborhood kernel
\begin{equation}
    \mathcal{K}_{jk} = \exp\!\left(-\frac{\|\bm{r}_j-\bm{r}_k\|^2}{2\sigma^2}\right),
\end{equation}
where $\sigma > 0$ controls the neighborhood width, the reference vectors are updated as
\begin{equation}
    \bm{w}_j \leftarrow \bm{w}_j + \eta \mathcal{K}_{j k_i^*}(\bm{x}_i - \bm{w}_j),
    \label{eq:kohonen_update}
\end{equation}
where $\eta \in (0,1)$ is the learning rate.

Batch SOM (BSOM) \cite{Kohonen1990} replaces this sequential update with
\begin{equation}
    \bm{w}_j = \frac{\sum_{i=1}^N \mathcal{K}_{j k_i^*}\bm{x}_i}{\sum_{i=1}^N \mathcal{K}_{j k_i^*}},
    \label{eq:batch_som_update}
\end{equation}
which updates all reference vectors with per-iteration computational complexity $O(NM+M^2)$. Although this is often summarized as $O(NM)$ when $M \ll N$, we retain the $M^2$ term explicitly because we also consider large-$M$ settings.

Both the sequential and batch forms are widely used for topographic mapping and often preserve neighborhood relations in practice, but they are defined algorithmically rather than as the exact minimization of a single explicit smooth optimization objective. In particular, because the BMU index changes discontinuously across Voronoi boundaries, the induced update field is not generally the gradient of a scalar potential~\cite{Erwin1992}. This limitation motivated subsequent optimization-based formulations of SOM-type learning.

\subsection{Objective-Based SOM Formulations and Their Limitations}

Following earlier Bayesian and distortion-based views of SOM-type learning \cite{Luttrell1994}, Heskes \cite{Heskes1999} formulated topographic mapping by the convolved distortion
\begin{equation}
    J_{\text{CD}}=\sum_{i=1}^N \sum_{j=1}^M p_{ij}\left(\sum_{k=1}^M h_{jk}\|\bm{x}_i-\bm{w}_k\|^2\right),
    \label{eq:convolved_distortion}
\end{equation}
where $p_{ij}\ge 0$, $\sum_{j=1}^M p_{ij}=1$, and $h_{jk}$ is typically given by the normalized Gaussian kernel
\begin{equation}
    h_{jk}=\frac{\exp\!\left(-\frac{\|\bm{r}_j-\bm{r}_k\|^2}{2\sigma^2}\right)}{\sum_{l=1}^M \exp\!\left(-\frac{\|\bm{r}_j-\bm{r}_l\|^2}{2\sigma^2}\right)}.
    \label{eq:normalized_kernel}
\end{equation}
This formulation incorporates neighborhood interaction directly into the objective, but it is linear in $p_{ij}$ and thus degenerates into hard assignments. In addition, the direct evaluation of \eqref{eq:convolved_distortion} incurs a per-iteration computational cost of $O(NM^2)$.

To obtain soft assignments, Graepel et al. \cite{Graepel1997} introduced Soft Topographic Vector Quantization (STVQ), which is based on the entropy-regularized objective
\begin{equation}
    J_{\text{STVQ}}=J_{\text{CD}}+\lambda \sum_{i=1}^N \sum_{j=1}^M p_{ij}\ln p_{ij},
    \label{eq:stvq_objective}
\end{equation}
where $\lambda>0$. This yields soft assignments of softmax form, but the neighborhood-coupled structure remains. To reduce the cost, Heskes \cite{Heskes2001} applied a bias-variance decomposition:
\begin{align}
    J_{\text{BVD}}
     & =\sum_{i=1}^N \sum_{j=1}^M p_{ij}\left(\|\bm{x}_i-\tilde{\bm{w}}_j\|^2+V_j(\mathcal{W})\right)\notag \\
     & \quad+\lambda \sum_{i=1}^N \sum_{j=1}^M p_{ij}\ln p_{ij},
    \label{eq:bvd_objective}
\end{align}
where
\begin{equation}
    \tilde{\bm{w}}_j=\sum_{k=1}^M h_{jk}\bm{w}_k,
    \quad
    V_j(\mathcal{W})=\sum_{k=1}^M h_{jk}\|\bm{w}_k-\tilde{\bm{w}}_j\|^2.
    \label{eq:bvd_defs}
\end{equation}
This reduces the per-iteration cost to $O(NM+M^2)$ under dense neighborhood interactions. Although this is more efficient than the original $O(NM^2)$ formulation, the $M^2$ term due to neighborhood coupling still remains, which can become a practical limitation when the number of nodes is large.

\section{Proposed Method: SOM-OLP}
\label{sec:proposed_method}

In this section, we propose Self-Organizing Maps with Optimized Latent Positions (SOM-OLP), an optimization-based topographic mapping method that introduces a continuous latent position for each data point. We derive a separable surrogate local cost from a continuous extension of the STVQ neighborhood distortion and formulate the SOM-OLP objective based on it.

\subsection{Continuous Latent Extension and Surrogate Local Cost}
\label{subsec:proposed_objective}

For a data point $\bm{x}_i\in\mathbb{R}^D$ and a latent node located at $\bm{r}_j\in\mathbb{R}^L$, the STVQ neighborhood distortion is given by
\begin{equation}
    E_i(\bm{r}_j)=\sum_{k=1}^M h_{jk}\|\bm{x}_i-\bm{w}_k\|^2,
    \label{eq:stvq_discrete_potential}
\end{equation}
where $\bm{w}_k\in\mathbb{R}^D$ is the reference vector of node $k$, and $h_{jk}$ is the normalized neighborhood function defined in \eqref{eq:normalized_kernel}. To extend this formulation to a continuous latent space, let \mbox{$\Omega\subset\mathbb{R}^L$} be a compact set containing $\mathcal{R}$, and define the continuous neighborhood function by
\begin{equation}
    h(\bm{r},\bm{r}_k)=\frac{\exp\!\left(-\frac{\|\bm{r}-\bm{r}_k\|^2}{2\sigma^2}\right)}{\sum_{l=1}^M \exp\!\left(-\frac{\|\bm{r}-\bm{r}_l\|^2}{2\sigma^2}\right)}.
    \label{eq:continuous_kernel}
\end{equation}
The resulting continuous neighborhood distortion is
\begin{equation}
    E_i(\bm{r})\coloneqq\sum_{k=1}^M h(\bm{r},\bm{r}_k)\|\bm{x}_i-\bm{w}_k\|^2.
    \label{eq:continuous_potential}
\end{equation}
Since $E_i$ is continuous on the compact set $\Omega$, it attains a minimum on $\Omega$. Intuitively, smaller values of $E_i(\bm{r})$ indicate latent positions that are more compatible with $\bm{x}_i$. We therefore define the latent position of $\bm{x}_i$ by
\begin{equation}
    \bm{v}_i\in\argmin_{\bm{r}\in\Omega}E_i(\bm{r}).
    \label{eq:latent_coordinate_definition}
\end{equation}
We then approximate $E_i$ locally around $\bm{v}_i$. Assuming that $\bm{v}_i$ is a local minimizer in the interior of $\Omega$, the first-order term vanishes at $\bm{v}_i$, and a local quadratic approximation gives
\begin{equation}
    E_i(\bm{r}_j)\approx E_i(\bm{v}_i)+\frac{1}{2}(\bm{v}_i-\bm{r}_j)^\top\mathbf{H}_i(\bm{v}_i-\bm{r}_j),
    \label{eq:quadratic_expansion_general}
\end{equation}
where $\mathbf{H}_i=\nabla^2 E_i(\bm{v}_i)$. Under a locally isotropic approximation $\mathbf{H}_i\approx 2\gamma\mathbf{I}$, with $\mathbf{I}\in\mathbb{R}^{L\times L}$ denoting the identity matrix and $\gamma\ge 0$, this becomes
\begin{equation}
    E_i(\bm{r}_j)\approx E_i(\bm{v}_i)+\gamma\|\bm{v}_i-\bm{r}_j\|^2.
    \label{eq:quadratic_expansion}
\end{equation}

Equation~\eqref{eq:quadratic_expansion} separates the local behavior of $E_i(\bm{r}_j)$ into a $j$-independent offset and a quadratic deviation term. In the no-neighborhood limit $h_{jk}\to\delta_{jk}$, \eqref{eq:stvq_discrete_potential} reduces to
\begin{equation}
    E_i(\bm{r}_j)\to\|\bm{x}_i-\bm{w}_j\|^2,
    \label{eq:no_neighbor_limit}
\end{equation}
which is the direct quantization error. Motivated by this limiting form together with the local quadratic approximation, we introduce the separable surrogate local cost
\begin{equation}
    \ell_{ij}\coloneqq\|\bm{x}_i-\bm{w}_j\|^2+\gamma\|\bm{v}_i-\bm{r}_j\|^2.
    \label{eq:local_cost}
\end{equation}
This surrogate preserves the direct data-space distortion term and introduces a latent-space proximity term.

\subsection{Optimization Problem}

Using \eqref{eq:local_cost}, we formulate SOM-OLP as the following constrained optimization problem:
\begin{equation}
    \begin{aligned}
        \underset{\mathbf{P},\mathcal{V},\mathcal{W}}{\text{minimize}} \quad & J_{\text{SOM-OLP}}                     \\
        \text{subject to}\quad                                               & p_{ij}\ge 0 \quad \forall i,\forall j, \\
                                                                             & \sum_{j=1}^M p_{ij}=1 \quad \forall i,
    \end{aligned}
    \label{eq:som_olp_problem}
\end{equation}
where
\begin{align}
    J_{\text{SOM-OLP}}
     & =\sum_{i=1}^N\sum_{j=1}^M p_{ij}\left(\|\bm{x}_i-\bm{w}_j\|^2+\gamma\|\bm{v}_i-\bm{r}_j\|^2\right)\notag \\
     & \quad+\lambda\sum_{i=1}^N\sum_{j=1}^M p_{ij}\ln p_{ij}.
    \label{eq:som_olp_objective}
\end{align}
Here, $\mathcal{V}=\{\bm{v}_i\}_{i=1}^N$ denotes the set of latent positions. Equation~\eqref{eq:som_olp_objective} is an entropy-regularized objective based on the separable surrogate local cost in \eqref{eq:local_cost}, and it can be evaluated in $O(NM)$ time per iteration.

The parameter $\lambda>0$ controls the strength of entropy regularization and hence the softness of the assignment weights, while $\gamma\ge 0$ controls the balance between the data-space distortion and the latent-space proximity term. When $\gamma=0$, SOM-OLP reduces to entropy-regularized fuzzy $c$-means~\cite{efcm}; when $\gamma=0$ and $\lambda\to 0$, it further reduces to standard \mbox{$k$-means}~\cite{kmeans}.

\subsection{Update Rules}
\label{subsec:optimization}

We optimize the problem in \eqref{eq:som_olp_problem} by block coordinate descent (BCD). Since $J_{\text{SOM-OLP}}$ is convex with respect to each block of variables, $\mathbf{P}$, $\mathcal{V}$, and $\mathcal{W}$, when the others are fixed, each update is obtained by minimizing $J_{\text{SOM-OLP}}$ with respect to the corresponding block.

\subsubsection{Update of $\mathbf{P}$}
With $\mathcal{V}$ and $\mathcal{W}$ fixed, minimizing $J_{\text{SOM-OLP}}$ by the method of Lagrange multipliers yields
\begin{equation}
    p_{ij}=\frac{\exp\!\left(-\frac{1}{\lambda}\left(\|\bm{x}_i-\bm{w}_j\|^2+\gamma\|\bm{v}_i-\bm{r}_j\|^2\right)\right)}
    {\sum_{k=1}^M \exp\!\left(-\frac{1}{\lambda}\left(\|\bm{x}_i-\bm{w}_k\|^2+\gamma\|\bm{v}_i-\bm{r}_k\|^2\right)\right)}.
    \label{eq:update_p_gamma}
\end{equation}
Thus, each data point is softly assigned by a softmax over a cost that combines data-space distortion and latent-space proximity.

\subsubsection{Update of $\mathcal{V}$}
With $\mathbf{P}$ and $\mathcal{W}$ fixed, the stationary condition $\nabla_{\bm{v}_i}J_{\text{SOM-OLP}}=\bm{0}$ yields
\begin{equation}
    \bm{v}_i=\sum_{j=1}^M p_{ij}\bm{r}_j.
    \label{eq:update_v}
\end{equation}
Hence, $\bm{v}_i$ is the weighted average of the node coordinates and provides a latent position for $\bm{x}_i$.

\subsubsection{Update of $\mathcal{W}$}
With $\mathbf{P}$ and $\mathcal{V}$ fixed, the stationary condition $\nabla_{\bm{w}_j}J_{\text{SOM-OLP}}=\bm{0}$ yields
\begin{equation}
    \bm{w}_j=\frac{\sum_{i=1}^N p_{ij}\bm{x}_i}{\sum_{i=1}^N p_{ij}}.
    \label{eq:update_w}
\end{equation}
Thus, each reference vector is updated as the weighted centroid of the data points assigned to it.

\subsection{Algorithm and Computational Complexity}
\label{subsec:algorithm_summary}

\begin{algorithm}[t]
    \caption{SOM-OLP}
    \label{alg:som_olp}
    \begin{algorithmic}[1]
        \setlength{\itemsep}{0.2em}
        \REQUIRE Data points $\mathcal{X}=\{\bm{x}_i\}_{i=1}^N$, node coordinates $\mathcal{R}=\{\bm{r}_j\}_{j=1}^M$, hyperparameters $\gamma,\lambda$, and tolerance $\varepsilon$
        \ENSURE $\mathcal{W},\mathcal{V},\mathbf{P}$
        \STATE Initialize $\mathcal{W}^{(0)}$ by PCA using \eqref{eq:pca_init_w}
        \STATE Initialize $\mathbf{P}^{(0)}$ using \eqref{eq:init_p}
        \REPEAT
        \STATE Update $\mathcal{V}$ using \eqref{eq:update_v}
        \STATE Update $\mathcal{W}$ using \eqref{eq:update_w}
        \STATE Update $\mathbf{P}$ using \eqref{eq:update_p_gamma}
        \UNTIL{the relative change in $J_{\text{SOM-OLP}}$ is at most $\varepsilon$}
        \RETURN $\mathcal{W},\mathcal{V},\mathbf{P}$
    \end{algorithmic}
\end{algorithm}

Algorithm~\ref{alg:som_olp} summarizes the cyclic BCD procedure of SOM-OLP. We first initialize the reference vectors $\mathcal{W}^{(0)}$ by a principal component analysis (PCA)-based scheme \cite{Jolliffe2002}, and then initialize the assignment probabilities $\mathbf{P}^{(0)}$ from $\mathcal{W}^{(0)}$.

Let $\tilde{\mathbf X}$ denote the centered version of $\mathbf X$, and let $\bm u_1,\dots,\bm u_K$ be the leading $K=\min\{L,D\}$ principal directions of $\tilde{\mathbf X}$. The node coordinates $\bm r_j$ are normalized along each axis to obtain $\tilde{\bm r}_j\in[-1,1]^K$, and let $\bm s=(s_1,\dots,s_K)$ denote the standard deviations of the projected data along these principal directions. The initial reference vector of node $j$ is then set as
\begin{equation}
    \bm w_j^{(0)}=\bm\mu+2\sum_{l=1}^K \tilde r_{jl}s_l\bm u_l,
    \label{eq:pca_init_w}
\end{equation}
where $\bm\mu$ is the data mean. Since the latent positions $\mathcal{V}$ are not available at initialization, we initialize the assignment probabilities using only the data-space distortion term\footnote{Random initialization of $\mathbf{P}^{(0)}$ also led to stable optimization in our preliminary experiments.}:
\begin{equation}
    p_{ij}^{(0)}=\frac{\exp\!\left(-\frac{1}{\lambda}\|\bm x_i-\bm w_j^{(0)}\|^2\right)}{\sum_{k=1}^M \exp\!\left(-\frac{1}{\lambda}\|\bm x_i-\bm w_k^{(0)}\|^2\right)}.
    \label{eq:init_p}
\end{equation}

After initialization, the variables are updated cyclically in the order $\mathcal{V}\rightarrow\mathcal{W}\rightarrow\mathbf{P}$. Under this cyclic BCD scheme, the objective value is monotonically non-increasing. Convergence can be determined by the relative change in the objective value. Specifically, letting $J^{(t)}$ denote the objective value at iteration $t$, the algorithm is regarded as converged when
\begin{equation}
    \frac{\left|J^{(t)}-J^{(t-1)}\right|}{\max\left\{1,\left|J^{(t-1)}\right|\right\}} \le \varepsilon,
    \label{eq:convergence}
\end{equation}
where $\varepsilon$ is a small positive constant, and the denominator prevents numerical instability when $\left|J^{(t-1)}\right|$ is close to zero.

Each update in Algorithm~\ref{alg:som_olp} has closed form, yielding $O(NM)$ per-iteration complexity for SOM-OLP, compared with $O(NM^2)$ for STVQ and $O(NM+M^2)$ for the BVD-based formulation under dense neighborhood interactions. 

\subsection{Interpretation of Topographic Consistency}
\label{subsec:topographic_bias}

Although $J_{\text{SOM-OLP}}$ contains no explicit node--node interaction term, substituting \eqref{eq:update_v} and \eqref{eq:update_w} into \eqref{eq:som_olp_objective} yields
\begin{align}
    \tilde{J}
     & =\frac{1}{2}\sum_{j=1}^M \sum_{i=1}^N \sum_{l=1}^N \frac{p_{ij}p_{lj}}{n_j}\|\bm{x}_i-\bm{x}_l\|^2\notag                                                 \\
     & \quad+\frac{\gamma}{2}\sum_{i=1}^N \sum_{j=1}^M \sum_{k=1}^M p_{ij}p_{ik}\|\bm{r}_j-\bm{r}_k\|^2+\lambda\sum_{i=1}^N\sum_{j=1}^M p_{ij}\ln p_{ij} \notag \\
     & =\operatorname{tr}(\mathbf X^\top \mathbf L_{\text{data}} \mathbf X)
    +\gamma\operatorname{tr}(\mathbf R^\top \mathbf L_{\text{lat}} \mathbf R)
    +\lambda\sum_{i=1}^N\sum_{j=1}^M p_{ij}\ln p_{ij},
    \label{eq:topographic_bias_rewrite}
\end{align}
where
\begin{align}
    \mathbf L_{\text{data}} &= \mathbf I_N-\mathbf P\mathbf D_n^{-1}\mathbf P^\top, \\
    \mathbf L_{\text{lat}}  &= \mathbf D_n-\mathbf P^\top\mathbf P, \\
    \mathbf D_n            &= \operatorname{diag}(n_1,\dots,n_M), \\
    n_j                    &= \sum_{i=1}^N p_{ij}.
\end{align}%

The first and second terms can be viewed as graph-Laplacian-type regularizers in the data space and latent space, respectively. For each node, the former favors assigning larger common weights to nearby data points than to distant ones. For each data point, the latter favors assigning larger weights to nearby latent nodes than to distant ones. Thus, the proposed objective encourages assignment structures consistent with neighborhood relations in both the data space and the latent space.

\section{Numerical Experiments}

\begin{table}[t]
    \centering
    \caption{Hyperparameter search space for each method.}
    \label{tab:hyperparam_search}
    \begin{tabular}{lccc}
        \hline
        Method  & Hyperparameter           & Search range       & Search scale \\
        \hline
        BSOM    & $\sigma$                 & $[10^{-3},10^{3}]$ & log          \\
        \hline
        STVQf   & $\sigma$                 & $[10^{-3},10^{3}]$ & log          \\
                & $\lambda$                & $[10^{-3},10^{3}]$ & log          \\
        \hline
        GTM     & $m$                      & $\{2,3,\dots,15\}$ & linear       \\
                & $s$                      & $[10^{-3},10^{3}]$ & log          \\
                & $\lambda_{\text{reg}}$ & $[10^{-3},10^{3}]$ & log          \\
        \hline
        SOM-OLP & $\gamma$                 & $[10^{-3},10^{3}]$ & log          \\
                & $\lambda$                & $[10^{-3},10^{3}]$ & log          \\
        \hline
    \end{tabular}
\end{table}

We compared Batch SOM (BSOM), Soft Topographic Vector Quantization (STVQ), its fast implementation based on bias--variance decomposition (denoted here by STVQf), Generative Topographic Mapping (GTM)~\cite{Bishop1998}, and the proposed SOM-OLP. GTM is a probabilistic generative model for topographic mapping; for this method, we used the \texttt{eGTM} class from the Python package \texttt{ugtm} (v2.0)~\cite{Gaspar2018}. All methods were initialized using PCA to eliminate variability due to random initialization. For all methods, the maximum number of iterations was set to 1000 and the convergence tolerance to $10^{-4}$. Unless otherwise noted, we used a two-dimensional square latent grid with $M=16^2$ nodes on $[-1,1]\times[-1,1]$.

As evaluation metrics, we used Trustworthiness (TW), Continuity (CN), and Quantization Error (QE). TW and CN assess two complementary aspects of neighborhood preservation between the data space and the latent space: TW measures the extent to which neighbors in the latent space are also neighbors in the original data space, whereas CN measures the extent to which neighbors in the original data space remain neighbors in the latent space~\cite{VennaKaski2001,LeeVerleysen2009}. In contrast, QE measures how accurately the data points are represented by the learned reference vectors in the data space~\cite{Polzlbauer2004}. For TW and CN, the number of neighbors was set to $5$.

For hyperparameter tuning, we used Optuna~\cite{optuna2019} with the multivariate Tree-structured Parzen Estimator (TPE) sampler. For each method and dataset, one Optuna study consisted of 100 trials maximizing $(\mathrm{TW}+\mathrm{CN})/2$. For the saddle-manifold and Digits case studies, we ran five Optuna studies with different sampler seeds for each method and selected the final model with the smallest QE among the five resulting models. The hyperparameter search ranges are listed in Table~\ref{tab:hyperparam_search}. All experiments were conducted on a Windows machine with an Intel Core i7-14700 CPU and 31.7\,GB RAM.


\subsection{Saddle Manifold}

The saddle-manifold data were generated by independently sampling $x_1$ and $x_2$ from the uniform distribution on $[-1,1]$, and then defining $x_3 = x_1^2 - x_2^2 + \xi$, where $\xi$ is Gaussian noise drawn from the normal distribution with mean $0$ and variance $0.1^2$. The number of data points was set to \mbox{$N=500$}. In this experiment, the generated three-dimensional data were used without additional normalization. For visualization in both the data space and the latent space, the data points were colored consistently according to their $x_3$ values so that the correspondence between the two spaces could be examined.

For the saddle-manifold dataset, the selected hyperparameters were $\sigma=0.1184$ for BSOM, $\sigma=0.0637$ and $\lambda=8.70\times10^{-3}$ for STVQf, $m=14$, $s=271.4$, and $\lambda_{\text{reg}}=1.35\times10^{-3}$ for GTM, and $\gamma=73.79$ and $\lambda=1.696$ for SOM-OLP.

\begin{table}[t]
    \caption{Quantitative comparison on the saddle dataset. \textbf{Bold}: best; \underline{underlined}: second best.}
    \label{tab:saddle_results_best_qe}
    \centering
    \begin{tabular}{lccccc}
        \hline
        Method  & TW $\uparrow$      & CN $\uparrow$      & QE $\downarrow$    & Iters $\downarrow$ & \shortstack{Time $\downarrow$ \\ {[ms/iteration]}} \\
        \hline
        BSOM    & 0.9945             & 0.9952             & 0.0089             & \textbf{17}        & \textbf{0.40}                 \\
        STVQf   & 0.9945             & 0.9952             & \textbf{0.0057}    & \underline{34}     & 4.09                          \\
        GTM     & \underline{0.9979} & \underline{0.9986} & \underline{0.0087} & 105                & 16.94                         \\
        SOM-OLP & \textbf{0.9986}    & \textbf{0.9992}    & 0.0102             & 110                & \underline{3.86}              \\
        \hline
    \end{tabular}
\end{table}

\begin{figure*}[t]
    \centering
    \includegraphics[width=0.92\textwidth]{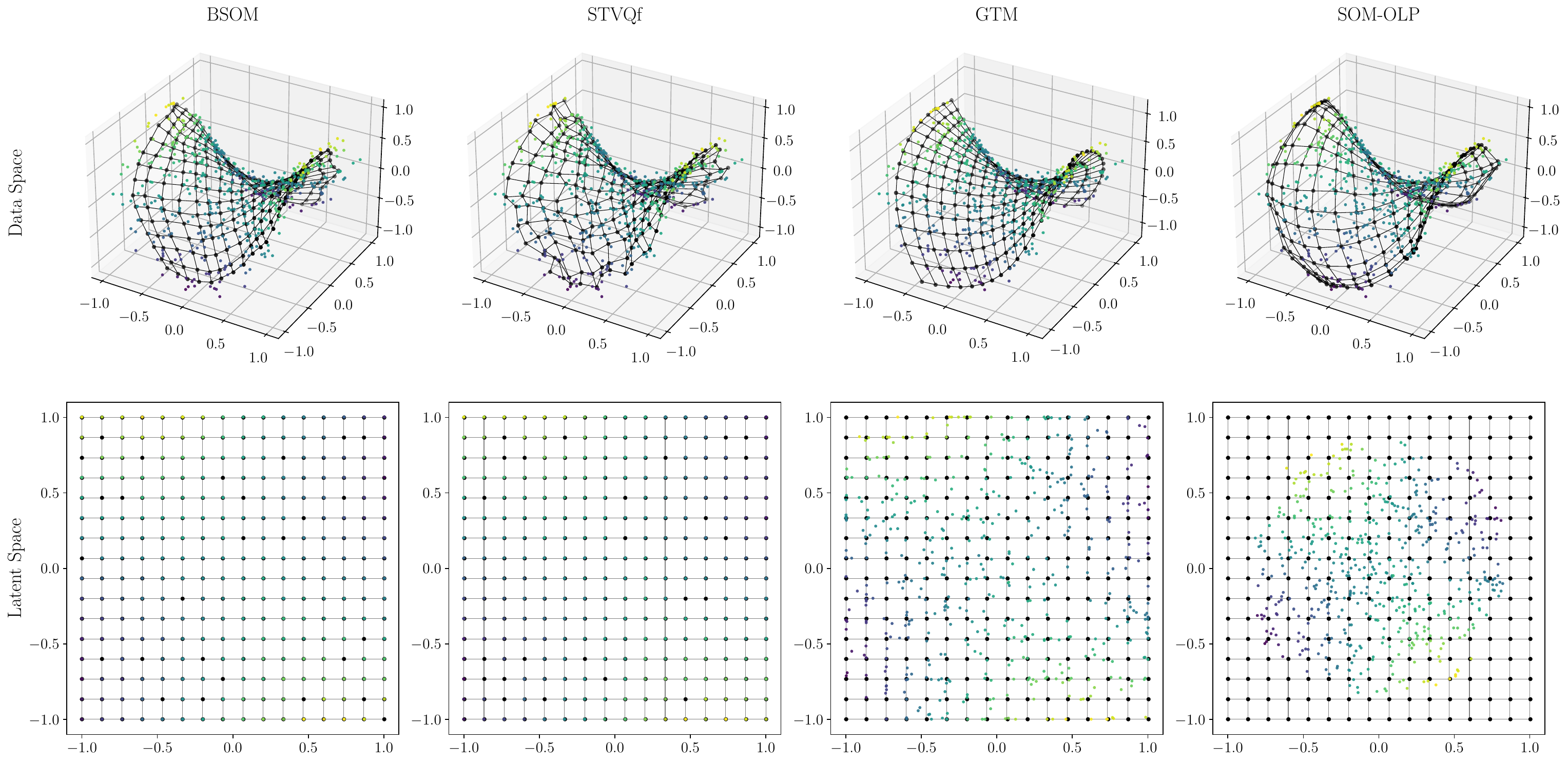}
    \caption{Comparison of BSOM, STVQf, GTM, and SOM-OLP on the saddle dataset. For each method, the data-space view shows the learned reference vectors, while the latent-space view shows the corresponding representation of the data points. The same color coding based on the original $x_3$ values is used in both spaces to facilitate visual comparison of neighborhood relationships across the two spaces.}
    \label{fig:compare_som_saddle_best_qe}
\end{figure*}

The quantitative results are reported in Table~\ref{tab:saddle_results_best_qe}. SOM-OLP achieved the best TW and CN values, whereas STVQf attained the smallest QE, and GTM showed the second-best performance in TW, CN, and QE. In terms of computational cost, BSOM required the fewest iterations and the smallest time per iteration. SOM-OLP required more iterations to converge, but its per-iteration time was smaller than those of STVQf and GTM under this experimental setting. These results suggest that SOM-OLP provides competitive topology-preserving performance on the saddle dataset while maintaining moderate computational cost among the compared objective-based methods.

Fig.~\ref{fig:compare_som_saddle_best_qe} shows that all four methods capture the overall structure of the saddle manifold in the data space. Since the selected models were obtained under a tuning criterion based on $(\mathrm{TW}+\mathrm{CN})/2$, and qualitative properties may vary with the hyperparameter setting, we do not overinterpret the visual differences. By construction, BSOM and STVQf provide discrete node-based latent representations, whereas GTM and SOM-OLP provide latent positions. Under the present setting, GTM appears to reflect the latent-node arrangement more strongly, whereas SOM-OLP yields a more flexible continuous embedding. Thus, the main qualitative observation is that SOM-OLP combines latent positions with competitive topology-preserving performance.

\subsection{Digits Dataset}

We evaluated the performance of each method using the Digits dataset from scikit-learn~\cite{sklearn}. This dataset comprises $N=1,797$ grayscale images of handwritten digits ($8 \times 8$ pixels), resulting in a $D=64$ dimensional data space. Each pixel value is an integer in the range $[0, 16]$. We normalized the input as $\mathbf{X} = \mathbf{X}_{\mathrm{raw}} / 16$ to scale the components within $[0, 1]$. For visualization, data points are colored according to their ground-truth labels ($0, \dots, 9$).

For the Digits dataset, the selected hyperparameters were $\sigma=0.0812$ for BSOM, $\sigma=0.0556$ and $\lambda=0.154$ for STVQf, $m=15$, $s=2.385$, and $\lambda_{\text{reg}}=0.558$ for GTM, and $\gamma=0.815$ and $\lambda=0.331$ for SOM-OLP.

\begin{table}[t]
    \caption{Quantitative comparison on the Digits dataset. \textbf{Bold}: best; \underline{underlined}: second best.}
    \label{tab:digits_results_best_qe}
    \centering
    \begin{tabular}{lccccc}
        \hline
        Method  & TW $\uparrow$      & CN $\uparrow$      & QE $\downarrow$    & Iters $\downarrow$ & \shortstack{Time $\downarrow$ \\ {[ms/iteration]}} \\
        \hline
        BSOM    & 0.9871             & 0.9657             & 1.2217             & \textbf{21}        & \textbf{11.45}                \\
        STVQf   & \underline{0.9872} & 0.9685             & \underline{0.9315} & \underline{33}     & \underline{25.96}             \\
        GTM     & \textbf{0.9910}    & \underline{0.9746} & 1.1518             & 89                 & 58.85                         \\
        SOM-OLP & 0.9864             & \textbf{0.9761}    & \textbf{0.8895}    & 37                 & 29.77                         \\
        \hline
    \end{tabular}
\end{table}

\begin{figure*}[t]
    \centering
    \includegraphics[width=0.95\linewidth]{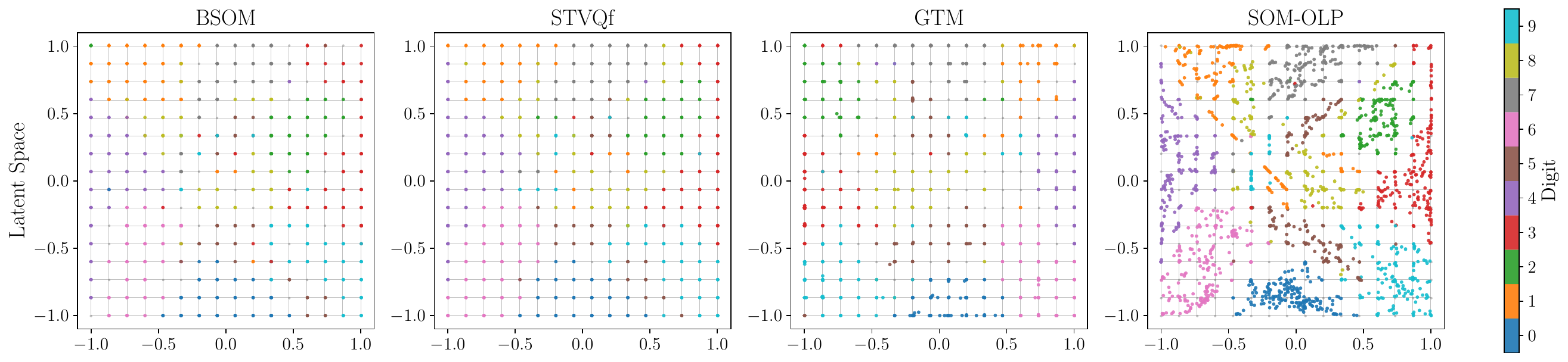}
    \caption{Latent representations of the Digits dataset. While BSOM and STVQf are constrained to discrete nodes, GTM and SOM-OLP provide continuous latent positions. Notably, SOM-OLP exhibits a more adaptive distribution that reflects the local density of the data manifold.}
    \label{fig:compare_som_digits_best_qe}
\end{figure*}

Table~\ref{tab:digits_results_best_qe} summarizes the quantitative metrics. GTM achieved the highest TW, whereas SOM-OLP outperformed the other methods in CN and QE, with STVQf also showing competitive performance. In terms of computational cost, BSOM was the most efficient in both iterations and time per iteration. Although SOM-OLP required more computation than BSOM and STVQf, it achieved the best CN and QE values while requiring approximately half the time per iteration of GTM. These results suggest that SOM-OLP provides competitive performance on the Digits dataset, with a favorable balance between mapping quality and computational cost.

Fig.~\ref{fig:compare_som_digits_best_qe} highlights structural differences in the learned mappings. BSOM and STVQf yield discrete representations in which data points are assigned to grid nodes, whereas GTM and SOM-OLP provide continuous latent positions. Under the present experimental conditions, GTM appears to produce a relatively grid-constrained distribution, although this tendency may depend on the hyperparameter setting, whereas SOM-OLP yields a more flexible representation that adapts to the local density of the digit clusters. Since the selected models were obtained under a tuning criterion based on $(\mathrm{TW}+\mathrm{CN})/2$, we do not overinterpret these visual differences. Rather, the main observation is that SOM-OLP provides flexible latent positions while maintaining competitive neighborhood-preserving performance, consistent with its favorable QE result.

\subsubsection{Scalability with Respect to the Number of Nodes}

\begin{table}[t]
    \centering
    \caption{Per-iteration wall-clock time (ms) versus grid size on the Digits dataset ($T=20$ iterations). OOM: out-of-memory error.}
    \label{tab:scalability_digits}
    \begin{tabular}{c|rrrr}
        \hline
        $M$                & BSOM                & STVQf           & GTM     & SOM-OLP            \\
        \hline
        $2\,500 = 50^2$    & \textbf{81}         & \underline{263} & 288     & 265                \\
        $10\,000 = 100^2$  & \textbf{359}        & 1\,173          & 1\,143  & \underline{1\,064} \\
        $22\,500 = 150^2$  & \textbf{975}        & 3\,247          & 3\,324  & \underline{2\,430} \\
        $40\,000 = 200^2$  & \textbf{4\,018}     & 10\,356         & 8\,268  & \underline{4\,641} \\
        $62\,500 = 250^2$  & \underline{22\,030} & 157\,937        & 55\,339 & \textbf{8\,617}    \\
        $90\,000 = 300^2$  & OOM                 & OOM             & OOM     & \textbf{12\,988}   \\
        $122\,500 = 350^2$ & OOM                 & OOM             & OOM     & \textbf{17\,509}   \\
        $160\,000 = 400^2$ & OOM                 & OOM             & OOM     & \textbf{23\,259}   \\
        $202\,500 = 450^2$ & OOM                 & OOM             & OOM     & \textbf{29\,897}   \\
        $250\,000 = 500^2$ & OOM                 & OOM             & OOM     & \textbf{36\,768}   \\
        \hline
    \end{tabular}
\end{table}

To evaluate scalability with respect to the number of nodes, we measured the per-iteration wall-clock time on the Digits dataset while increasing the latent grid size from $M=50^2$ to $500^2$. For each method, the hyperparameters were fixed to those selected in the preceding Digits experiment. For each grid size, the runtime was averaged over $T=20$ iterations. The results are summarized in Table~\ref{tab:scalability_digits}.

For relatively small to moderate grids ($M \le 40{,}000$), BSOM is the fastest method, and SOM-OLP is consistently the second fastest. As the grid size increases further, however, the relative advantage of SOM-OLP becomes more apparent. At $M=62{,}500$, SOM-OLP achieves the smallest per-iteration time among all compared methods. Moreover, for $M \ge 90{,}000$, BSOM, STVQf, and GTM all run out of memory in our experimental environment, whereas SOM-OLP remains executable up to $M=250{,}000$.

These empirical results are consistent with the computational structure of the methods. BSOM and STVQf involve neighborhood-dependent computations, and GTM also becomes increasingly costly as the number of latent nodes grows in the present experiment. In contrast, SOM-OLP is based on a separable local cost and avoids explicit node--node convolutions. As a result, SOM-OLP scales more favorably as $M$ increases, although BSOM remains more efficient at smaller node counts.

\subsection{MNIST Dataset}

\begin{figure}[t]
    \centering
    \includegraphics[width=0.72\linewidth]{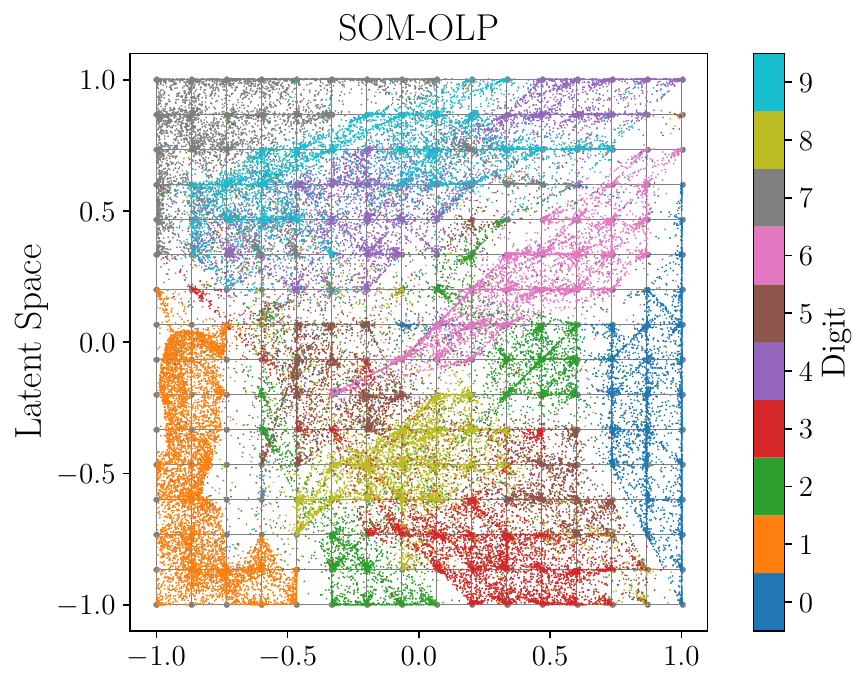}
    \caption{Latent representation of the MNIST dataset obtained by SOM-OLP with a $16\times16$ grid. Data points are colored by their ground-truth digit labels.}
    \label{fig:mnist_full}
\end{figure}

To further examine the scalability of SOM-OLP, we applied it to the MNIST handwritten digit dataset~\cite{LeCun1998}, which consists of $N=70{,}000$ grayscale images ($28\times28$ pixels) in a $D=784$ dimensional space with 10 classes. Pixel values were normalized to $[0,1]$ by dividing by 255. The hyperparameters were fixed to those selected in the $N=2{,}000$ subsampled experiment ($\gamma=2.767$, $\lambda=2.439$), and SOM-OLP was trained on the full dataset and a $16\times16$ grid ($M=256$). Training converged in 36 iterations, with approximately 5.9\,s per iteration. Fig.~\ref{fig:mnist_full} shows the resulting latent representation. Despite the large scale and high dimensionality of the dataset, SOM-OLP organizes the digit classes into broadly separated regions in the latent space. At the same time, noticeable overlap remains among some classes, suggesting that a fixed two-dimensional latent grid cannot fully disentangle all class relationships in MNIST. These results indicate that SOM-OLP scales to large high-dimensional datasets while still providing a meaningful topographic organization.

\subsection{Real-world Datasets}

To further evaluate generality, we compared the methods on 16 benchmark datasets, summarized in Table~\ref{tab:uci_combined}. As a simple linear baseline, we also included PCA with a two-dimensional projection. Although PCA does not explicitly preserve topology, it provides a standard reference for assessing whether topology-aware nonlinear methods offer practical advantages over a widely used linear dimensionality-reduction method. The datasets were taken from scikit-learn~\cite{sklearn} and the UCI Machine Learning Repository~\cite{UCI}. Missing values, when present, were imputed by the feature-wise median, and all features were standardized to zero mean and unit variance.

\begin{table*}[t]
    \caption{$(\mathrm{TW}+\mathrm{CN})/2$ on benchmark datasets ($M=16\times16$, standardized). \textbf{Bold}: best; \underline{underlined}: second best.}
    \label{tab:uci_combined}
    \centering
    \begin{tabular}{lrrccccc}
        \hline
        Dataset       & $N$   & $D$ & PCA    & BSOM               & STVQf              & GTM                & SOM-OLP            \\
        \hline
        Iris          & 150   & 4   & 0.9820 & 0.9799             & \underline{0.9838} & 0.9787             & \textbf{0.9869}    \\
        Wine          & 178   & 13  & 0.9041 & 0.9503             & 0.9531             & \textbf{0.9595}    & \underline{0.9593} \\
        Sonar         & 208   & 60  & 0.8754 & 0.9277             & \underline{0.9391} & \textbf{0.9395}    & 0.9344             \\
        Seeds         & 210   & 7   & 0.9615 & 0.9730             & \underline{0.9751} & 0.9700             & \textbf{0.9806}    \\
        Glass         & 214   & 9   & 0.8884 & \underline{0.9527} & 0.9490             & \textbf{0.9574}    & 0.9497             \\
        Heart Disease & 303   & 13  & 0.8348 & 0.9222             & 0.9265             & \underline{0.9317} & \textbf{0.9350}    \\
        Ecoli         & 336   & 7   & 0.9284 & 0.9663             & \textbf{0.9694}    & 0.9621             & \underline{0.9671} \\
        Ionosphere    & 351   & 34  & 0.8866 & \underline{0.9143} & 0.9078             & \underline{0.9143} & \textbf{0.9151}    \\
        Dermatology   & 366   & 34  & 0.8982 & 0.9452             & 0.9556             & \textbf{0.9616}    & \underline{0.9607} \\
        Breast Cancer (WDBC)          & 569   & 30  & 0.9137 & 0.9434             & \underline{0.9451} & 0.9428             & \textbf{0.9517}    \\
        Banknote Authentication      & 1372  & 4   & 0.9835 & 0.9939             & 0.9928             & \underline{0.9948} & \textbf{0.9963}    \\
        Yeast         & 1484  & 8   & 0.8791 & \underline{0.9560} & \underline{0.9560} & 0.9436             & \textbf{0.9600}    \\
        Digits        & 1797  & 64  & 0.8846 & 0.9717             & 0.9747             & \textbf{0.9768}    & \underline{0.9766} \\
        Image Segmentation  & 2310  & 19  & 0.9167 & 0.9867             & 0.9852             & \textbf{0.9877}    & \underline{0.9870} \\
        Rice (Cammeo and Osmancik)          & 3810  & 7   & 0.9610 & 0.9876             & 0.9883             & \underline{0.9891} & \textbf{0.9911}    \\
        Dry Bean      & 13611 & 16  & 0.9637 & 0.9830             & 0.9824             & \underline{0.9831} & \textbf{0.9865}    \\
        \hline
    \end{tabular}
\end{table*}

\begin{figure}[t]
    \centering
    \includegraphics[width=0.8\linewidth]{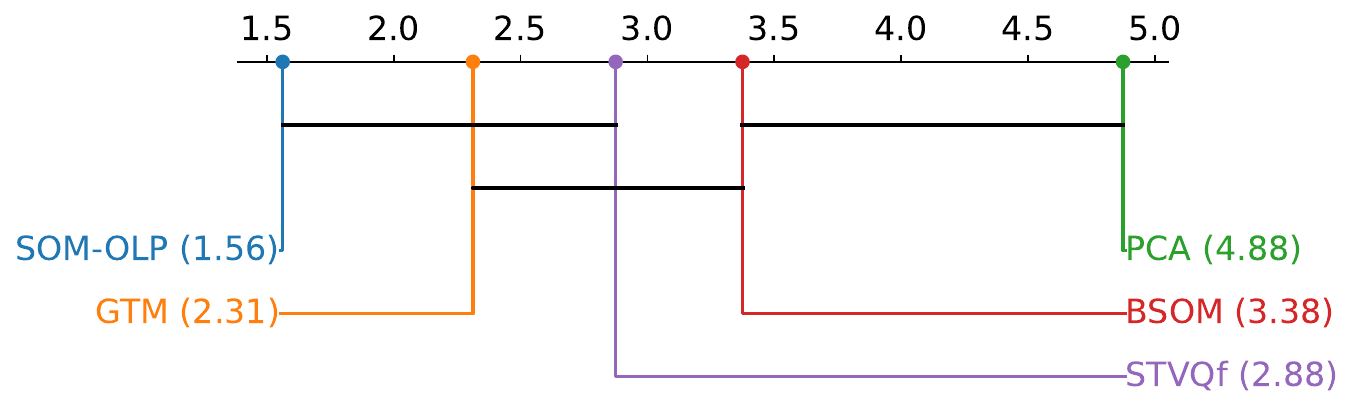}
    \caption{Critical-difference diagram based on the average ranks of $(\mathrm{TW}+\mathrm{CN})/2$ over the 16 benchmark datasets.}
    \label{fig:compare_som_uci_cd}
\end{figure}

Table~\ref{tab:uci_combined} summarizes the results on the 16 datasets. SOM-OLP achieved the best score on 9 datasets and the second-best score on 5 datasets, indicating the most consistently strong performance overall. GTM attained the best score on 6 datasets, whereas STVQf was best on Ecoli, and BSOM did not attain the best score on any dataset. PCA was consistently inferior to the nonlinear methods, which confirms the benefit of topology-aware nonlinear mapping.

The average-rank analysis led to the same overall conclusion. SOM-OLP achieved the best average rank (1.56), followed by GTM (2.31), STVQf (2.88), BSOM (3.38), and PCA (4.88). The Friedman test rejected the null hypothesis that all methods perform equivalently ($\chi^2=39.75$, $p=4.88\times10^{-8}$). In the critical-difference diagram shown in Fig.~\ref{fig:compare_som_uci_cd}, methods connected by a horizontal bar do not differ significantly at the $0.05$ level. The Nemenyi post-hoc analysis showed that SOM-OLP significantly outperformed BSOM, whereas its differences from STVQf and GTM were not significant. PCA performed significantly worse than STVQf, GTM, and SOM-OLP.

\section{Discussion}
\label{sec:discussion}

Among related methods, GTM \cite{Bishop1998} and Parametrized SOM (PSOM) \cite{Walter1996} are closely related in that they also relax the strictly discrete node-based representation of classical SOM. However, their formulations differ from that of SOM-OLP. GTM provides continuous latent representations through a probabilistic generative model, whereas PSOM achieves continuity by interpolating a node-supported manifold. In contrast, SOM-OLP introduces a latent position for each data point directly into the objective function and optimizes it jointly with the assignment weights and reference vectors. This yields a simpler reference-vector-based formulation and clarifies its relation to entropy-regularized fuzzy $c$-means and $k$-means.

Other recent directions, such as SOM-VAE \cite{Fortuin2019}, SatSOM~\cite{Urbanik2025}, and Topological Autoencoders \cite{Moor2020}, address different goals. SOM-VAE focuses on deep discrete representation learning, with a particular emphasis on time series; SatSOM addresses plasticity control in continual learning; and Topological Autoencoders preserve topological structures of the input space through persistent-homology-based regularization. By contrast, SOM-OLP targets batch topographic mapping in a shallow reference-vector-based framework, with local topographic consistency induced by the objective itself.

Overall, SOM-OLP can be positioned as an objective-based topographic mapping method with continuous latent positions, closed-form updates, monotonic non-increase under cyclic BCD, and $O(NM)$ per-iteration complexity without explicit $O(M^2)$ node--node interactions. Moreover, it includes entropy-regularized fuzzy $c$-means as a special case and further reduces to $k$-means.

\section{Conclusion}
\label{sec:conclusion}

This paper presented SOM-OLP, an objective-based topographic mapping method that introduces a continuous latent position for each data point. Starting from the neighborhood distortion of STVQ, we constructed a separable surrogate local cost and derived an entropy-regularized objective with closed-form cyclic updates for assignment probabilities, latent positions, and reference vectors. As a result, SOM-OLP retains linear per-iteration complexity while avoiding the explicit node--node coupling that appears in previous objective-based SOM formulations. Experiments on a synthetic saddle manifold, scalability studies on the Digits and MNIST datasets, and 16 benchmark datasets showed that SOM-OLP achieves strong neighborhood preservation and quantization performance, favorable scalability for large numbers of latent nodes and large datasets, and the best average rank among the compared methods on the benchmark datasets. Future work includes broader comparisons with manifold learning and clustering methods, further analysis of the surrogate formulation and its theoretical properties, and more systematic strategies for hyperparameter selection.

\section*{Acknowledgment}
The authors used ChatGPT (OpenAI) to improve the wording and readability of portions of this manuscript.


\end{document}